\documentclass[10pt,final,conference,twocolumn]{IEEEtran}
\usepackage{amsmath}
\usepackage{graphicx}
\usepackage[hidelinks]{hyperref}
\title{A Measure of Similarity of Textual Data Using Spearman's Rank Correlation Coefficient}

\author{\IEEEauthorblockN{Nino Arsov\IEEEauthorrefmark{1}, Milan Dukovski\IEEEauthorrefmark{2}, Blagoja Evkoski\IEEEauthorrefmark{3} and Stefan Cvetkovski\IEEEauthorrefmark{4} }
\IEEEauthorblockA{Faculty of Computer Science and Engineering,\\
Ss. Cyril and Methodius University\\
Skopje, Macedonia\\
Email: 
\IEEEauthorrefmark{1}arsov.nino@students.finki.ukim.mk
\IEEEauthorrefmark{2}dukovski.milan@students.finki.ukim.mk\\
\IEEEauthorrefmark{3}evkoski.blagoja@students.finki.ukim.mk
\IEEEauthorrefmark{4}cvetkovski.stefan@students.finki.ukim.mk
}
}

\begin{document}
\maketitle
\begin{abstract}
In the last decade, many diverse advances have occurred in the field of information extraction from data. Information extraction in its simplest form takes place in computing environments, where structured data can be extracted through a series of queries.
The continuous expansion of quantities of data have therefore provided an opportunity for knowledge extraction (KE) from a textual document (TD). A typical problem of this kind is the extraction of common characteristics and knowledge from a group of TDs, with the possibility to group such similar TDs in a process known as \textit{clustering}.

In this paper we present a technique for such KE among a group of TDs related to the common characteristics and meaning of their content. Our technique is based on the Spearman's Rank Correlation Coefficient (SRCC), for which the conducted experiments have proven to be comprehensive measure to achieve a high-quality KE.

\end{abstract}
\begin{IEEEkeywords}
Spearman's Rank Correlation Coefficient, textual data, similarity, knowledge extraction, vector space
\end{IEEEkeywords}

\section{Introduction}

Over the past few years, the term \textit{big data} has become an important key point for research into data mining and information retrieval. Through the years, the quantity of data managed across enterprises has evolved from a simple and imperceptible task to an extent to which it has become the central performance improvement problem. In other words, it evolved to be the next frontier for innovation, competition and productivity \cite{manyika2011big}.
Extracting knowledge from data is now a very competitive environment. Many companies process vast amounts of customer/user data in order to improve the \textit{quality of experience (QoE)} of their customers. For instance, a typical use-case scenario would be a book seller that performs an automatic extraction of the content of the books a customer has bought, and subsequently extracts knowledge of what customers prefer to read. The knowledge extracted could then be used to recommend other books. Book recommending systems are typical examples where data mining techniques should be considered as the primary tool for making future decisions \cite{xue2010research}.

KE from TDs is an essential field of research in data mining and it certainly requires techniques that are reliable and accurate in order to neutralize (or even eliminate) uncertainty in future decisions. Grouping TDs based on their content and mutual key information is referred to as \textit{clustering}. Clustering is mostly performed with respect to a measure of similarity between TDs, which must be represented as vectors in a \textit{vector space} beforehand \cite{manning2008introduction}. News aggregation engines can be considered as a typical representative where such techniques are extensively applied as a sub-field of natural language processing (NLP).

In this paper we present a new technique for measuring similarity between TDs, represented in a vector space, based on SRCC - "a statistical measure of association between two things" \cite{spearman1904proof}, which in this case things refer to TDs. The mathematical properties of SRCC (such as the ability to detect nonlinear correlation) make it compelling to be researched into. Our motivation is to provide a new technique of improving the quality of KE based on the well-known association measure SRCC, as opposed to other well-known TD similarity measures.

The paper is organized as follows: Section~\ref{sec:background} gives a brief overview of the vector space representation of a TD and the corresponding similarity measures, in Section~\ref{sec:related} we address conducted research of the role of SRCC in data mining and trend prediction. Section~\ref{sec:solution} is a detailed description of the proposed technique, and later, in Section~\ref{sec:experiments} we present clustering and classification experiments conducted on several sets of TDs, while Section~\ref{sec:conclusion} summarizes our research and contribution to the broad area of statistical text analysis.

\section{Background}
\label{sec:background}

In this section we provide a brief background of vector space representation of TDs and existing similarity measures that have been widely used in statistical text analysis. To begin with, we consider the representation of documents. 

\subsection{Document Representation}
A document $d$ can be defined as a finite sequence of terms (independent textual entities within a document, for example, words), namely $d=(t_1,t_2,\dots ,t_n)$. A general idea is to associate \textit{weight} to each term $t_i$ within $d$, such that
$$d=(w_{t_1}, w_{t_2}, \dots ,w_{t_n}),$$
which has proven superior in prior extensive research \cite{salton1988term}. The most common weight measure is \textit{Term Frequency - Inverse Document Frequency (TF-IDF)}. TF is the frequency of a term within a single document, and IDF represents the importance, or uniqueness of a term within a set of documents $D=\{d_1, d_2, \dots ,d_m\}$. TF-IDF is defined as follows:
$$TF\mbox{-}IDF_{t,d}=TF_{t,d}\cdot IDF_{t,D},$$ where
$$TF_{t,d}=\frac{f(t,d)}{n}  \quad IDF_{t,D}=\log \frac{|D|}{|\{d\in D : t\in d\}|},$$
such that $f$ is the number of occurrences of $t$ in $d$ and $\log$ is used to avoid very small values close to zero.

Having these measures defined, it becomes obvious that each $w_i$, for $i=1,\dots ,n$ is assigned the TF-IDF value of the corresponding term. It turns out that each document is represented as a vector of TF-IDF weights within a vector space model (VSM) with its properties \cite{turney2010frequency}.

\subsection{Measures of Similarity}
Different ways of computing the similarity of two vector exist. There are two main approaches in similarity computation:
\begin{itemize}
\item[$1.$]\textit{Deterministic} - similarity measures exploiting algebraic properties of vectors and their geometrical interpretation. These include, for instance, cosine similarity (CS), Jaccard coefficients (for binary representations), etc.
\item[$2.$]\textit{Stochastic} - similarity measures in which uncertainty is taken into account. These include, for instance, statistics such as Pearson's Correlation Coefficient (PCC) \cite{pearson1895note}.
\end{itemize}
Let $\mathbf{u}$ and $\mathbf{v}$ be the vector representations of two documents $d_1$ and $d_2$. Cosine similarity simply measures $cos\theta$, where $\theta$ is the angle between $\mathbf{u}$ and $\mathbf{v}$

\begin{itemize}
\item[](cosine similarity) $$cs(\mathbf{u}, \mathbf{v})=\frac{\mathbf{u} \cdot \mathbf{v}}{\|\mathbf{u}\|\|\mathbf{v}\|}$$
\item[](PCC) $$r=\frac{1}{n-1} \sum_{i=1}^{n} \frac{u_i - \bar{u}}{s_u}\frac{v_i-\bar{v}}{s_v},$$
 where $$s_u=\sqrt{\frac{1}{n-1}\sum_{i=1}^{n}(u_i-\bar{u})^{2}}$$
\end{itemize}

All of the above measures are widely used and have proven efficient, but an important aspect is the lack of importance of the order of terms in textual data. It is easy for one to conclude that, two documents containing a single sentence each, but in a reverse order of terms, most deterministic methods fail to express that these are actually very similar. On the other hand, PCC detects only linear correlation, which constraints the diversity present in textual data. In the following section, we study relevant research in solving this problem, and then in Sections~\ref{sec:solution} and~\ref{sec:experiments} we present our solution and results.

\section{Related Work}
\label{sec:related}

A significant number of similarity measures have been proposed and this topic has been thoroughly elaborated. Its main application is considered to be clustering and classification of textual data organized in TDs. In this section, we provide an overview of relevant research on this topic, to which we can later compare our proposed technique for computing vector similarity.

KE (also referred to as \textit{knowledge discovery}) techniques are used to extract information from unstructured data, which can be subsequently used for applying supervised or unsupervised learning techniques, such as clustering and classification of the content \cite{rajman1998text}. Text clustering should address several challenges such as vast amounts of data, very high dimensionality of more than 10,000 terms (dimensions), and most importantly - an understandable description of the clusters \cite{beil2002frequent}, which essentially implies the demand for high quality of extracted information.

Regarding high quality KE and information accuracy, much effort has been put into improving similarity measurements. An improvement based on linear algebra, known as Singular Value Decomposition (SVD), is oriented towards word similarity, but instead, its main application is document similarity \cite{deerwester1990indexing}. Alluring is the fact that this measure takes the advantage of synonym recognition and has been used to achieve human-level scores on multiple-choice synonym questions from the Test of English as a Foreign Language (TOEFL) in a technique known as Latent Semantic Analysis (LSA) \cite{landauer1997solution} \cite{turney2010frequency}. 

Other semantic term similarity measures have been also proposed, based on information exclusively derived from large corpora of words, such as Pointwise Mutual Information (PMI), which has been reported to have achieved a large degree of correctness in the synonym questions in the TOEFL and SAT tests \cite{mihalcea2006corpus}. 

Moreover, normalized \textit{knowledge-based measures}, such as Leacock \& Chodrow \cite{leacock1998combining}, Lesk (\textit{"how to tell a pine cone from an ice-cream cone"} \cite{lesk1986automatic}, 	or measures for the depth of two concepts (preferably vebs) in the Word-Net taxonomy \cite{wu1994verbs} have experimentally proven to be efficient. Their accuracy converges to approximately 69\%, Leacock \& Chodrow and Lesk have showed the highest precision, and having them combined turns out to be the approximately optimal solution \cite{mihalcea2006corpus}.

\section{The Spearman's Rank Correlation Coefficient Similarity Measure}
\label{sec:solution}

The main idea behind our proposed technique is to introduce uncertainty in the calculations of the similarity between TDs represented in a vector space model, based on the nonlinear properties of SRCC. Unlike PCC, which is only able to detect linear correlation, SRCC's nonlinear ability provides a convenient way of taking different ordering of terms into account.

\subsection*{Spearman's Rank Correlation Coefficient}

The Spreaman's Rank Correlation Coefficient \cite{spearman1904proof}, denoted $\rho$, has a from which is very similar to PCC. Namely, for $n$ raw scores $U_i, V_i$ for $i=1,\dots ,n$ denoting TF-IDF values for two document vectors $\mathbf{U}, \mathbf{V}$, 
\pagebreak
$$\rho = 1 - \frac{6\sum_{i=1}^{n}(u_i-v_i)^{2}}{n(n^{2}-1)},$$
where $u_i$ and $v_i$ are the corresponding ranks of $U_i$ and $V_i$, for $i=0,\dots ,n-1$. A metric to assign the ranks of each of the TF-IDF values has to be determined beforehand. Each $U_i$ is assigned a rank value $u_i$, such that $u_i=0,1,\dots ,n-1$. It is important to note that the metric by which the TF-IDF values are ranked is essentially their sorting criteria. 
A convenient way of determining this criteria when dealing with TF-IDF values, which emphasize the \textit{importance} of a term within a TD set, is to sort these values in an ascending order. Thus, the largest (or most important) TF-IDF value within a TD vector is assigned the rank value of $n-1$, and the least important is assigned a value of $0$.

\subsubsection*{An Illustration of the Ranking TF-IDF Vectors}

Consider two TDs $d_1$ and $d_2$, each containing a single sentence.\\\\
\textbf{Document 1:} John had asked Mary to marry him before she left.\\
\textbf{Document 2:} Before she left, Mary was asked by John to be his wife.\\\\
Now consider these sentences lemmatized:\\\\
\textbf{Document 1:} John have ask Mary marry before leave.\\
\textbf{Document 2:} Before leave Mary ask John his wife.\\\\

Let us now represent $d_1$ and $d_2$ as TF-IDF vectors for the vocabulary in our small corpus.

$$d_1=
\begin{bmatrix}
 0&0&0.09&0&0&0&0.09&0&0\\
\end{bmatrix}
$$
$$d_2=
\begin{bmatrix}
 0&0&0&0.09&0&0&0&0&0.09\\
\end{bmatrix}$$

\begin{table}[h]
\centering
\begin{tabular}{ll}

\hline
\textbf{CS}($d_1$,$d_2$) & \textbf{SRCC}($d_1$,$d_2$) \\ \hline
0.00             & -0.285714         \\ \hline
\end{tabular}
\vspace{1mm}
\caption{A comparison between the CS and SRCC similarity measures.}
\label{tab:1}
\end{table}

The results in Table~\ref{tab:1} show that SRCC performs much better in knowledge extraction. The two documents' contents contain the same idea expressed by terms in a different order that John had asked Mary to marry him before she left. It is obvious that \textit{cosine similarity} cannot recognize this association, but SRCC has successfully recognized it and produced a similarity value of -0.285714.

SRCC is essentially conducive to semantic similarity. Rising the importance of a term in a TD will eventually rise its importance in another TD. But if the two TDs are of different size, the terms' importance values will also differ, by which a nonlinear association will emerge. This association will not be recognized by PCC at all (as it only detects linear association), but SRCC will definitely catch this detail and produce the desirable similarity value. The idea is to use SRCC to catch such terms which drive the semantic context of a TD, which will follow a nonlinear and lie on a polynomial curve, and not on the line $x=y$.

In our approach, we use a non-standard measure of similarity in textual data with simple and common frequency values, such as TF-IDF, in contrast to the statement that simple frequencies are not enough for high-quality knowledge extraction \cite{turney2010frequency}. In the next section, we will present our experiments and discuss the results we have obtained.


\section{Experiments}
\label{sec:experiments}

In order to test our proposed approach, we have conducted a series of experiments. In this section, we briefly discuss the outcome and provide a clear view of whether our approach is suitable for knowledge extraction from textual data in a semantic context.

We have used a dataset of 14 TDs to conduct our experiments. There are several subjects on which their content is based: (\textit{aliens, stories, law, news}) \cite{textfiles}. 

\subsection{Comparison Between Similarity Measures}
In this part, we have compared the similarity values produced by each of the similarity measures CS, SRCC and PCC. We have picked a few notable results and they are summarized in Table~\ref{tab:2} below.\\

\begin{table}[h]
\centering
\begin{tabular}{llll}
\hline
documents & 
$CS$           & $SRCC$          & $PCC$         \\
\hline\\
$d_1,d_4$   & \textbf{0.27}             & 0.13              & 0.22            \\
$d_1,d_5$   & 0.36             & 0.0018            & \textbf{0.05} \\
$d_1,d_8$   & \textbf{0.012} & 0.0063            & 0.05            \\
$d_6,d_{10}$  & 0.009            & \textbf{0.022}  & 0.01            \\
$d_2,d_4$   & \textbf{0.24}  & 0.16              & 0.2             \\
$d_7,d_{12}$  & 0.0073           & \textbf{0.0250} & 0.01            \\
$d_8,d_9$   & 0.97             & 0.95              & 0.97    \\
\hline    
\hspace{2mm}
\end{tabular}

\caption{A comparison between the measures CS, SRCC, ~PCC}
\label{tab:2}
\end{table}

In Table~\ref{tab:2} that SRCC mostly differs from CS and PCC, which also differ in some cases.For instance, $d_1$ refers to \textit{leadership in the nineties}, while $d_5$ refers to \textit{the family and medical lead act of 1993}. We have empirically observed that the general topics discussed in these two textual documents are very different. Namely, discusses different frameworks for leadership empowerment, while $d_5$ discusses medical treatment and self-care of employees. We have observed that the term \textit{employee} is the only connection between $d_1$ and $d_5$. The similarity value of CS of 0.36 is very unreal in this case, while PCC (0.05), and especially SRCC (0.0018) provide a much more realistic view of the semantic knowledge aggregated in these documents. Another example are $d_8$ and $d_9$. The contents of these documents are very straightforward and very similar, because they discuss \textit{aliens} seen by Boeing-747 pilots and $d_9$ discusses angels that were considered to be aliens. It is obvious that SRCC is able to detect this association as good as CS and PCC which are very good in such straightforward cases.

We have observed that SRCC does not perform worse than any other of these similarity measures. It does not always produce the most suitable similarity value, but it indeed does perform at least equally good as other measures. The values in Table~\ref{tab:2} are very small, and suggest that SRCC performs well in extracting tiny associations in such cases. It is mostly a few times larger than CS and PCC when there actually exist associations between the documents.

\begin{figure}[h]
\centering
\includegraphics[scale=0.7]{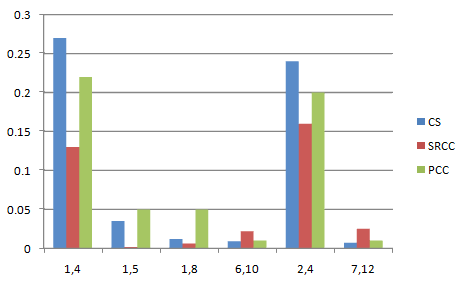}

\caption{A visual comparison of similarities produced by CS, SRCC and PCC}\label{fig:1}
\end{figure}

These results are visually summarized in Figure~\ref{fig:1}. The two above-described examples can be clearly seen as standing out.

\subsection{Non-linearity of Documents}
In this part we will briefly present the nonlinear association between some of the TDs we have used in our experiments. Our purpose is to point out that $(d_6,d_{10})$ and $(d_7,d_{12})$ are the pairs where SRCC is the most appropriate measure for the observed content, and as such, it is able to detect the nonlinear association between them. This can be seen in Figure~\ref{fig:2} below. The straightforward case of $d_8$ and $d_9$ also stands out here (SRCC can also detect it very well).

\begin{figure}[h]
\centering
\includegraphics[scale=0.5]{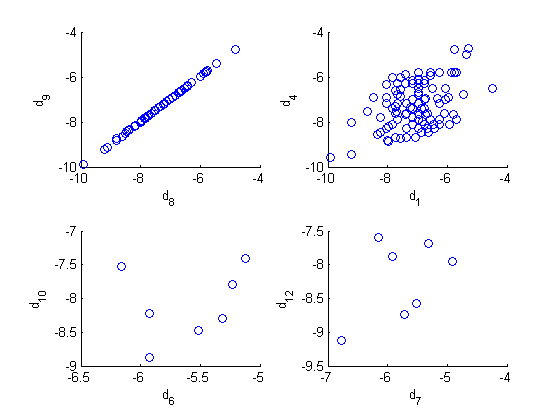}

\caption{The association between documents}\label{fig:2}
\end{figure}
The obtained results showed that our technique shows good performance on similarity computing, although it is not a perfect measure. But, it sure comes close to convenient and widely used similarity measures such as CS and PCC. The next section provides a conclusion of our research and suggestions for further work.
\section{Conclusion and Future Work}
\label{sec:conclusion}
In this paper we have presented a non-standard technique for computing the similarity between TF-IDF vectors. We have propagated our idea and contributed a portion of new knowledge in this field of text analysis. We have proposed a technique that is widely used in similar fields, and our goal is to provide starting information to other researches in this area. We consider our observations promising and they should be extensively researched.

Our experiments have proved that our technique should be a subject for further research. Our future work will concentrate on the implementation of machine learning techniques, such as clustering and subsequent classification of textual data. We expect an information of good quality to be extracted. To summarize, the rapidly emerging area of \textit{big data} and information retrieval is where our technique should reside and where it should be applied. 

\bibliographystyle{IEEEtran}
\bibliography{ref}

\end{document}